\documentclass[10pt,letterpaper]{article}

\usepackage{cogsci}

\cogscifinalcopy % Uncomment this line for the final submission 

\usepackage{pslatex}
\usepackage{apacite}
\usepackage{float} % Roger Levy added this and changed figure/table
\usepackage{graphicx} 
\usepackage{subcaption}
\usepackage{multicol}
\usepackage{multirow}
\usepackage{makecell}

\title{A Factorized Probabilistic Model of the Semantics of Vague Temporal Adverbials Relative to Different Events}

\author{{\large \bf Svenja Kenneweg (skenneweg@techfak.uni-bielefeld.de)} \\
  Bielefeld University, Faculty of Technology, Universitätsstraße 25, 33615 Bielefeld, Germany
  \AND {\large \bf Jörg Deigmöller (Joerg.Deigmoeller@honda-ri.de)} \\
  Honda Research Institute Europe, Carl-Legien-Straße 30, 63073 Offenbach am Main, Germany
  \AND {\large \bf Julian Eggert (Julian.Eggert@honda-ri.de)} \\
  Honda Research Institute Europe, Carl-Legien-Straße 30, 63073 Offenbach am Main, Germany
  \AND {\large \bf Philipp Cimiano (cimiano@techfak.uni-bielefeld.de)} \\
  Bielefeld University, Faculty of Technology, Universitätsstraße 25, 33615 Bielefeld, Germany}
\begin{document}

\maketitle

\begin{abstract}
% Include no author information in the initial submission, to facilitate
% blind review.  The abstract should be one paragraph, indented 1/8~inch on both sides,
% in 9~point font with single spacing. The heading ``{\bf Abstract}''
% should be 10~point, bold, centered, with one line of space below
% it. This one-paragraph abstract section is required only for standard
% six page proceedings papers. Following the abstract should be a blank
% line, followed by the header ``{\bf Keywords:}'' and a list of
% descriptive keywords separated by semicolons, all in 9~point font, as
% shown below.

Vague temporal adverbials, such as ”recently," "just" and ”long time ago,” describe the temporal distance between a past event and the utterance time, but leave the exact duration underspecified. 
In this paper, we introduce a factorized model that captures the semantics of these adverbials as probabilistic distributions. These distributions are composed with event-specific distributions to yield a contextualized meaning for an adverbial applied to a specific event. 
We fit the model's parameters using existing data capturing judgements of native speakers regarding the applicability of these vague temporal adverbials to events that took place a given time ago. Comparing our approach to a non-factorized model based on a single Gaussian distribution for each pair of event and temporal adverbial, we find out that, while both models have similar predictive power, our model is preferable in terms of Occam’s razor, as it is simpler and has a better extendability.

\end{abstract}

\section{Introduction}
Vague temporal adverbials such as \emph{recently}, \emph{just}, \emph{some time ago}, or \emph{long time ago} are used to describe the temporal recency or distance between an event that took place in the past and the utterance time. These temporal adverbials are vague in the sense that they do not specify the exact length of the temporal interval passed between the event in question and the utterance time, leaving this length under-specified. Yet, this length is not arbitrary, as corroborated by the following examples. 

While we can say \emph{`I just celebrated my birthday'} when it was a few days ago, we would rather not utter such a sentence when our birthday was months or years ago. 
Take another example \emph{`I just brushed my teeth'}. Such a sentence can be uttered when the event of brushing our teeth is minutes, but not days or weeks, ago. \\
Although both examples use the same vague temporal adverbial \emph{just}, the length of the temporal interval between the events and the utterance time varies, ranging from a few minutes to several days. This suggests that this length is shaped not only by the choice of the temporal adverbial, but also by the event itself. In fact, it has been argued that there are two crucial variables of an event that affect this length: duration and frequency \cite{jaarsveldschreuder}.

These insights indicate that a successful approach to modeling the semantics of vague temporal adverbials must be compositional, capturing not only how these adverbials define a temporal frame of reference but also how different events shape the resulting interpretation. \\
In this paper we present such a compositional approach to modeling the semantics of vague temporal adverbials. Grounded in the theory that the semantics of vague expressions can be viewed as having crisp yet uncertain boundaries \cite{LAWRY20091539, Vagueness_Williamson}, we represent the vague temporal adverbials and their corresponding events as probabilistic distributions. This allows us to assign higher probabilities to those timepoints that are most likely to be described by a given vague temporal adverbial and corresponding event. As it would not be realistic to assume that each vague temporal adverbial, such as \emph{just}, has one distinct meaning for each event, we factorize the interpretation: the meaning of an adverbial distribution remains fixed, and the event properties shape how that distribution is applied. \\
Therefore our compositional model relies on two types of probability density functions $P_{Adv}$ and $P_{Ev}$. When applied to an event that has happened $t$ times unites ago in the past, $P_{Ev}(t)$ transforms the distance between the $\texttt{reference\_time}-t$ into a distribution specific for the event $\textit{Ev}$, representing the probability that this event, which occurred $t$ time units ago, can actually be said to have happened \emph{before} the \texttt{reference\_time}. The \texttt{reference\_time} is the temporal anchor point relative to which the occurrence of the event is evaluated. 
The probability density function $P_{Adv}$ takes the result of the transformation by $P_{Ev}$ so that the distribution $P_{Adv}(P_{Ev}(t))$ then models the likelihood that the vague temporal adverbial $\textit{Adv}$ can be felicitously used to describe an event $\textit{Ev}$ that happened $t$ time units ago. 
 
The main contribution of our paper is the factorized compositional model to capturing the meaning of vague temporal adverbials, which is described in the Section "Compositional Model"\ref{sec:compositonalModel}. We fit the parameters of the model relying on data from  \citeA{kenneweg-etal-2024-empirical}, in which we elicited the judgments of English speakers regarding the applicability of the different vague temporal adverbials to describing events that are a particular time unit away from the utterance time. We compare our approach to a non-factorized approach as baseline that models the likelihood for every pair of adverb and event explicitly in terms of data fit (accuracy) and extendability.

\section{Background}

Vague predicates are characterized by the fact that there are borderline cases, that is cases where it is not clear whether the vague predicate applies or not. For example, the adjective \emph{tall} can be said to be vague as there are cases of people that can neither be clearly identified as \emph{tall} nor \emph{not tall}. 
Philosophically, vagueness has been characterized by reference to Sorites Paradox, which (instantiated for the case of tallness) would claim that anyone who is  merely 1mm shorter than a tall person intuitively counts as tall. Yet, if we start with a 2.5m person who is clearly tall, this principle leads us, by a series of small steps, to the absurd conclusion that a person who is only 10mm high would also count as tall.\\

% In many such cases, the judgments of whether the vague predicate applies is highly contextual, so that modeling the specific context and the interplay with the vague predicate has been an important research question. Take the adjective \emph{big} as an example. Whether something can be described as \emph{big} varies contextually depending on whether we are talking about, e.g., elephants or mice. 
% Our work is specifically concerned with capturing the context to which a vague temporal predicate applies, in this paper specifically for the case of temporal vague adverbs. 

While vagueness occurs across parts-of-speech, much of the literature has focused on adjectives \cite{Kamp2016, Kennedy, Lassiter, Solt2012}. \citeA{Kamp2016} have proposed a categorization of vague adjectives into \emph{relative} (e.g., tall, short) and \emph{absolute} adjectives (e.g., clean, dirty, straight).
%The interpretation of relative adjectives, also known as gradable adjectives \cite{Solt2012} depends on a comparison class \cite{Katz1967, Kamp2016, Solt2012}. 
Beyond adjectives, other parts-of-speech, like quantifiers (\emph{many}, \emph{few}) and nouns (\emph{mountain}) also exhibit vagueness. 
% All three categories exhibit vagueness through phenomena such as the existence of borderline cases (for which it is not clear whether they truly apply or not), the context-dependent meaning of the adjectives and the sorites paradox. Relative adjectives, also known as gradable adjectives \cite{Solt2012}, describe qualities that can vary in intensity or degree. Their interpretation depends on a comparison class \cite{Katz1967, Kamp2016, Solt2012}. Beyond adjectives, other parts-of-speech also represent cases of vague predicates. Take the case of quantifiers such as \emph{many} or \emph{few} which are vague with respect to the cardinality of the set they apply to and which is very context dependent. Another example are nouns; consider the case of the noun \emph{mountain}: At what point does a hill become a mountain? The Sorites Paradox applies here as well—does adding one meter to a hill’s height suddenly make it a mountain? 
In this paper, we consider the case of vague temporal adverbials such as \emph{recently}, \emph{just}, \emph{some time ago}, and \emph{long time ago}, which so far have not received substantial attention in the literature on vagueness. While we focus on this particular case, our compositional approach could nevertheless be generalized to model vagueness compositionally in general, for instance, to cases of spatial vagueness.

There are two prominent, fundamentally different philosophical frameworks of how to treat vagueness. The \emph{Epistemic theory} \cite{Vagueness_Williamson} assumes that vague predicates have precise but unknown boundaries, and that vagueness arises due to a lack of knowledge rather than inherent indeterminacy in the world itself. It posits that a predicate either applies or does not apply to a specific case, following a binary distinction with a certain probability. This probability depends on the degree of epistemic certainty about the threshold, which can be modeled using probability theory. In this context, the epistemic probability quantifies, e.g., the uncertainty that a specific individual’s height falls above or below the unknown threshold for being classified as "tall." As the available information increases, the epistemic probability can shift, reflecting greater confidence in determining whether a person qualifies as tall or not (for a more formal account, see \citeA{Climenhaga}). Therefore, the epistemic probability captures the agent's belief about the truth of the predicate, given the evidence at hand. This approach emphasizes that the vagueness is not an inherent feature of the concept itself, but rather a result of limited knowledge about the exact cutoff. Following the idea that predicates have crisp boundaries, \emph{Supervaluationism theory}, as proposed by \citeA{Fine1975}, uses a logical approach and models subjectivity in vague predicates by assuming that their extension varies across models. A person may be tall in some models of a theory, but not-tall in others. This leads to the definition of statements that are \emph{supertrue} (true in all models) and \emph{superfalse} (false in all models). While the \emph{Supervaluationism theory} captures logical subjectivity, it does not address where boundaries lie or how speakers determine predicate applicability in specific contexts.

In contrast to the approaches that assume that boundaries are crisp but unknown, there are approaches that attempt to capture the degree to which a vague predicate applies to a certain case. The most prominent approach along these lines, \emph{Fuzzy Set Theory} \cite{ZADEH1965338}, allows for different degrees of membership of an instance in a \emph{fuzzy set} along the closed interval $[0,1]$. A taller person might thus be member of the fuzzy set of \emph{`tall persons'} to a degree of 0.8. 

Along these lines, \emph{Prototype Theory} is a further approach that models the degree of membership with respect to some prototypical instance of the vague predicate \cite{prototypetheory}.  According to this theory, some instances of a concept are more representative than others; the most representative instances are the prototypes. The membership degree of an instance can then be defined by a similarity function to the prototype of the corresponding concept \cite{gradedmembership}. Consequently, the probability of an object belonging to a concept decreases with the distance to the prototype of this concept. This relationship can be modeled with a Gaussian distribution \cite{gaussconceptualspaces}. The \emph{conceptual spaces} theory by \citeA{conceptualspaces} also allows for the modeling of vague concepts in terms of convex geometrical regions in a metric space that allows to define the closeness of instances to the concept \cite{vaguenessgraded}.  
%The prototype of a concept can be defined using the metric of a conceptual space: It is a distinct sub-region, with concepts represented by regions defined by dimensions like hue, saturation and brightness \cite{conceptualspaces}. Voronoi diagrams can visualize concept divisions \cite{Douven2010} and vagueness arises at their boundaries where the distinctions between concepts becomes less defined.
% prototypes are sub-regions that have distinguished positions withing the region corresponding to the concept. 
% In the conceptual spaces framework, developed by \citeA{conceptualspaces}, concepts are represented geometrically, as regions in spaces that are defined in terms of one or more  representational dimensions. For example, humans represent color using a three-dimensional conceptual space defined in terms of the dimensions \emph{hue}, \emph{saturation}, and \emph{brightness}. Voronoi diagrams can be used to visualize how concepts can be divided into regions based on the proximity to prototypes \cite{Douven2010}. Vagueness then arises at the boundaries of these Voronoi diagrams where the distinctions between concepts becomes less defined.

Independently of which approach is followed, an important property of vague predicates is that their meaning is context-dependent. For instance, a baby is considered \emph{young} at the age of a few months but not at a few years, whereas an adult could still be considered \emph{young} at the age of 20 \cite{DAMERAU1977}. According to \citeA{Kamp_Keenan_1975}, particularly gradable adjectives like \emph{big} or \emph{small} should be interpreted relative to a reference class. 

Prior work has emphasized the necessity of a reference or comparison class in the interpretation of vague predicates, particularly adjectival ones. \citeA{Heim} discusses how adjectives access a reference class through their degree argument (DegP). For instance, in the sentence "Every girl is taller than 4 feet," the phrase "than 4 feet" provides the comparison standard against which the adjective taller is evaluated. \citeA{Schmidt2010} propose two distinct types of models for the interpretation of \emph{tall}: a threshold-based model, where an object is considered tall if it exceeds a contextually defined threshold, and a category-based model, where objects are assigned to clusters based on shared statistical properties such as height. Similarly to the threshold-based model, \citeA{Kennedy} posits a context-dependent "standard of comparison" which relies on a reference class. In "big dog" the reference class is "dog" and in "big children" "children". This reference class sets the threshold above which a dog or children is considered big and can be determined by using typical or average values of the reference class. \citeA{Qing2014} and \citeA{Lassiter2017} have introduced probabilistic models that capture how speakers and listeners pragmatically reason about vague predicates in relation to comparison classes.

While these approaches offer valuable insights into the context-sensitivity of vague adjectives, they do not adopt a fully compositional perspective in the sense of Frege's principle of compositionality \cite{Frege1953}, where complex meanings are systematically built from the meanings of their parts and their mode of combination. Moreover, in these accounts, the reference class is typically treated as a statistical distribution over a population of instances, rather than as something structured in its interaction with a specific event or object.

In the context of vague temporal adverbials such as we consider in this paper, the reference class relates to the different events that they can apply to. Previous work has identified that key parameters of events determining in how far they can be described by a certain vague temporal adverbial are the characteristic \emph{frequency} and \emph{duration} of the event \cite{jaarsveldschreuder}. This suggests that events that have a similar frequency and duration would provide a similar or comparable reference class. 

Building on this insight, we model the reference class by modeling the \emph{characteristic temporal signature} of events and making it comparable so that events with a similar frequency and duration are closer in the temporal signature space. The meaning of a vague temporal adverbial applied to a specific event results then from the compositional application of its specific signature to the \emph{characteristic temporal signature} of the event. 

Thus, our work extends previous probabilistic accounts of vagueness by introducing a compositional semantic structure that explicitly models how the meaning of vague expressions arises from the structured interaction between events and adverbials. This offers a novel framework for modeling vagueness not merely at the level of context sensitivity, but at the level of compositional meaning construction.

\section{Compositional Model}\label{sec:compositonalModel}

The model we introduce is a factorized model, following Frege's principle of compositionality \cite{Frege1953}, paraphrased as \emph{``The meaning of a complex predicate can be modeled via the meaning of its parts and how they are composed together''}. In our framework, these parts consist of the vague temporal adverbial (e.g. \emph{just}, \emph{recently},...) on the one hand and their reference class - the event (e.g. \emph{Brushing Teeth}, \emph{Sabbatical}, ...) on the other.

In our case of modeling the complex compositional meaning of a vague temporal adverbial and event, there are at least three important 'parts' to consider:

\begin{itemize}
\item Each temporal adverbial (e.g. \emph{just}, \emph{recently}, \emph{long time ago}) conveys a distinct notion of temporal proximity - thus having a distinct meaning 
\item Events can be categorized based on their \emph{characteristic temporal signature}, which is determined by factors such as duration and frequency \cite{jaarsveldschreuder}
\item The number $t$ of time units that have passed since the specific event happened until the time of utterance (\texttt{reference\_time})
\end{itemize}

% The model we propose in our paper is a factorized model that follows the principle of compositionality as proposed by Frege in the sense that the meaning of a complex predicate can be modeled via the meaning of its parts and how they are composed together \cite{Frege1953}. 

% In the case of modeling the meaning of vague temporal adverbials, there are at least three important elements or `parts' to consider:

% \begin{itemize}
% \item meaning of the vague temporal adverb as such, which differs for different adverbs such as just, recently, long time ago, etc. 
% \item the reference class, in our case corresponding to the event type to which the vague temporal predicate is applied
% \item the number $t$ of time units that have passed since the specific event happened until the time of utterance \emph{now}
% \end{itemize}

As we follow the \emph{epistemic theory} proposed by \citeA{Vagueness_Williamson}, we take the application of a particular vague temporal adverbial \emph{Adv} to an event \emph{Ev} that was $t$ time units ago to have crisp but unknown boundaries. We model the uncertainty in the boundaries using probability theory, and propose a factorized probabilistic model that relies on two probability density functions: $P_{\textit{Adv}}$ and $P_{\textit{Ev}}$. $P_{\textit{Adv}}$ is specific for each adverbial $\textit{Adv}$ and $P_{\textit{Ev}}$ for each event $\textit{Ev}$. 

The factorized model, which returns the probability that the adverbial $\textit{Adv}$ applies to an event $\textit{Ev}$ that happened $t$ time units ago, is as follows:

\[ P_{Adv}(P_{Ev}(t)) \]

% As we follow the epistemic theory proposed by \citeA{Vagueness_Williamson}, we take the application of a particular vague temporal adverbial \emph{Adv} to an event of type \emph{Ev} that was $t$ time units ago to have crisp but unknown boundaries that we model via probabilistic density functions. In particular, we propose a factorized model that relies on two probability density functions $P_{\textit{Adv}}$ and $P_{\textit{Ev}}$ that are specific for each adverbial $\textit{Adv}$ and event type $\textit{Ev}$. 

% The probability that the adverbial $\textit{Adv}$ applies to an event of $\textit{Ev}$ that happened $t$ time units ago is returned by the following factorized probability distribution:

% $P_{Adv}(P_{Ev}(t))$

% In our factorized compositional model, $P_{Ev}$ is a function that models the reference and is specific for the given event type $\textit{Ev}$. 
Intuitively, the (probability) function $P_{Ev}$ models the probability that an event $\textit{Ev}$ that happened $t$ time units \emph{before} the \texttt{reference\_time} can actually be said to have happened \emph{before} the \texttt{reference\_time}. Take the example of a teeth brushing event: a teeth brushing event that happened a few minutes ago has, from a probabilistic point of view, clearly happened before now (as now is here the reference\_time), while a teeth brushing event that happened a few milliseconds ago did less clearly happen \textit{before} \texttt{now}. The probabilistic distribution $P_{\textit{Ev}}$ thus captures the temporal signature of events \textit{Ev} with respect to the likelihood that we are willing to describe instances of the event as having happened \emph{before} depending on how long ago they actually happened. This event specific probability distribution models the reference class, as different events will generally have a different signature of this function. The function models the uncertainty in describing that one event \textit{Ev} occurs before some other time point or interval. 

Assuming that there is some uncertainty for each event in terms of when it took place exactly, we model events as Gaussian distributions of the following form:

\[
\frac{1}{\sqrt{2 \pi \sigma^2}} \exp\left(-\frac{1}{2} \frac{(x - \mu)^2}{\sigma^2}\right)
\]

Conversely, the uncertainty related to whether an event \textit{Ev} took place before another event or time interval can also be modeled as a Gaussian distribution as follows, where $erf$ is the Gauss error function:

\[\frac{1}{2}\left(erf\left(\frac{\mu}{\sqrt{2}\sigma}\right)+1\right) \]

We thus model the function \( P_{\textit{Ev}} \) as a Gaussian distribution of the above type, where \( \mu \) represents the number $t$ of time units that have passed since the specific event occurred. The parameter \( \sigma \) varies depending on the event, influencing the function to output higher probabilities for time units that are closer to or farther away from the utterance time.
%In conclusion the event-specific function $P_{Ev}$ maps the time units that have passed since the specific event instance happened into the normalized interval $[0,1]$.
% The event-specific probabilistic density functions map a temporal interval of $t$ time units that has passed since the specific event instance happened into a normalized interval $[0,1]$.
The second function of our model, the adverbial-specific function, $P_{Adv}$, takes this normalized value as input. As mentioned before, we follow \citeA{gaussconceptualspaces} and model each adverbial as a Gaussian distribution around its prototype. 
%Specifically, we use a normalized Gaussian distribution, which guarantees that the output of each adverbial-specific function, $P_{Adv}$, which is the probability of each adverbial, remains within 0 and 1, ensuring that they are consistent and interpretable. 
Thus, the adverbial-specific function $P_{Adv}$ is expressed as follow: 

\[\exp\left(-\frac{1}{2} \left(\frac{x - \mu_a}{\sigma_a}\right)^2\right) \]

This function has the two parameters $\mu_a$ and $\sigma_a$, which are the mean and standard deviation of the Gaussian distribution and defined individually for each adverbial.

Thus, the \emph{event-specific function} $P_{Ev}$ feeds into a normalized Gaussian distribution with specific parameters for each temporal adverbial. This final step yields a probability representing how likely it is that a human speaker would use a certain temporal adverbial, based on the probability of the described event's temporal precedence.

\section{Model Fitting}

We fit the parameters of the model using data from our previous study that elicited judgements of native speakers regarding the applicability of a vague temporal adverbial to an event that took place $t$ time units ago \cite{kenneweg-etal-2024-empirical}. Through an online survey, we collected probabilities reflecting the likelihood of a specific temporal adverbial for a given event and elapsed time since the event occurred. For this purpose, our participants, which were all adult native English speakers, rated the applicability of a temporal adverbial applied in english sentences to a past event on a Likert Scale. An example sentence from the survey is: “Tom’s Birthday was 1 day ago. Statement: Tom had his birthday recently.”

The study comprised 6 events (\emph{Brushing Teeth}, \emph{Birthday}, \emph{Vacation}, \emph{Sabbatical}, \emph{Year Abroad}, \emph{Marriage}) and 4 vague temporal adverbials (\emph{just}, \emph{recently}, \emph{some time ago}, \emph{long time ago}). For each event - adverbial pair, we explored its applicability across 7 elapsed times at which the event occurred. We divided the survey into 6 separate questionnaires, each targeting one event applied to all adverbials. Each questionnaire was completed by 100 participants recruited via the Prolific platform\footnote{https://www.prolific.com/}.
% \footnote{\href{https://www.prolific.com/}{https://www.prolific.com/}}.

In order to fit the parameters of our model, we normalized their Likert-scale responses into the interval $[0,1]$. This results in 16.800 data points (6 events x 4 adverbials x 7 times x 100 votes). Given the compositional nature of our model, we solve the combined function $P_{Adv}(P_{Ev}(t))$ for each event-adverbial pair, but over all data points - therefore for example finding the best parameters of \emph{Just} over all evaluated events. For instance, for the event \emph{Vacation}, this involved solving equations like $P_{Just}(P_{Vacation}(1 Month)) = 0.2$. Time values were standardized by converting them to minutes, with $1Month$ corresponding to $43800$ minutes (based on an average month length of 30,4 days). For fitting, we used least-squares optimization to minimize the residuals, i.e., the difference between the observed and modeled probabilities.   
% Initial guesses
% vagueAdverbials_initialguessmeans = {
%     "recently": 0.0,
%     "just": 0.0,
%     "some time ago": 0.8,
%     "long time ago": 1.0
% }
% initialguess_stds = [0.2] * len(vagueAdverbials_initialguessmeans)
% initialguess_stdRelationFunction = max_time_ago / 3

\begin{figure*}
\begin{center}
\includegraphics[width=\linewidth]{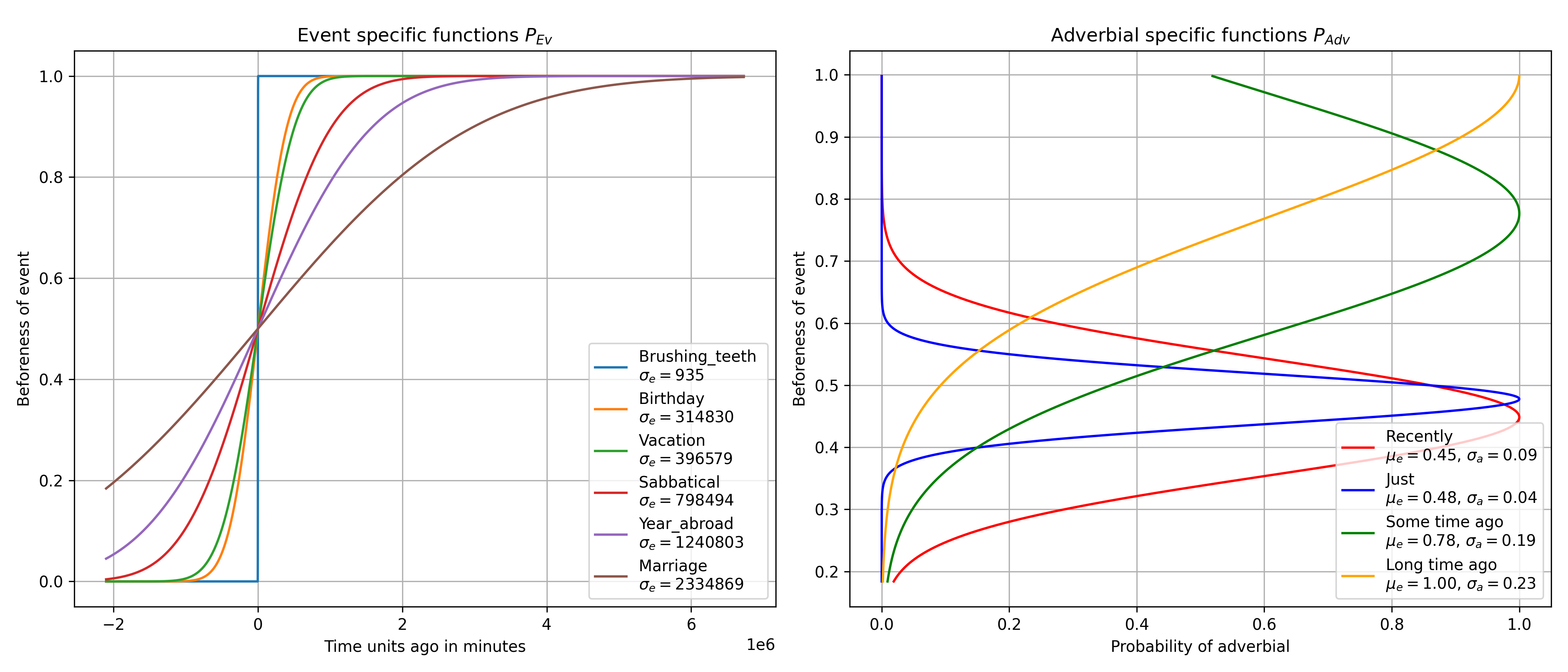}
\end{center}
\caption{Event-specific functions and adverbial-specific functions for all fitted events and adverbials.} 
\label{plot:fittingvacationalladverbials}
\end{figure*}

From the fitting we get the optimal parameters (means and standard deviations) for each adverbial-specific function, $P_{Adv}$, as well as the standard deviation for each event-specific function $P_{Ev}$ for all 4 fitted adverbials and 6 events.
The final standard deviations for each event-specific function $P_{Ev}$ are shown in Table \ref{tab:fittingresults_event}, while the means and standard deviations for the Gaussian distributions of each adverbial-specific function $P_{Adv}$ are presented in Table \ref{tab:fittingresults_adverbials}. To illustrate how these functions combine, Figure \ref{plot:fittingvacationalladverbials} shows the composition of all 6 event-specific functions with the 4 adverbial-specific functions.

\begin{table}[H]
\begin{center} 
\caption{Standard deviation of each event-specific function in minutes} 
\label{tab:fittingresults_event} 
\vskip 0.12in
\begin{tabular}{ll} 
\hline
Event    &  $\sigma_e$ \\
\hline
Brushing Teeth    &   935 \\
Birthday          &   314830 \\
Vacation          &   396579 \\
Sabbatical        &   798494 \\
Year Abroad       &   1240803 \\
Marriage          &   2334869 \\
\hline
\end{tabular} 
\end{center} 
\end{table}

\begin{table}[H]
\begin{center} 
\caption{Standard deviation and mean of each adverbial-specific function} 
\label{tab:fittingresults_adverbials} 
\vskip 0.12in
\begin{tabular}{lll} 
\hline
Adverbial    &  $\sigma_a$  & $\mu_a$\\
\hline
Just           &   0.04 & 0.48\\
Recently       &   0.09 & 0.45\\
Some Time Ago  &   0.19 & 0.78\\
Long Time Ago  &   0.23 & 1.00\\
\hline
\end{tabular} 
\end{center} 
\end{table}

\section{Evaluation}
We evaluate our compositional model $P_{Adv}(P_{Ev}(t))$ by comparing it with a baseline, a non-factorized model that relies on a single Gaussian distribution for each pair of vague temporal adverbial and event. We compare the two models in terms of two key aspects: \textbf{accuracy} and \textbf{extendability}. In our case, the non-factorized model is a normalized Gaussian function of the following form:

\[\exp\left(-\frac{1}{2} \left(\frac{x - \mu}{\sigma}\right)^2\right)\]

The Gaussian for the non-factorized model is fitted onto the same empirical data from \citeA{kenneweg-etal-2024-empirical} via least squares optimization, but with one difference: each event-adverbial pair is fitted its own Gaussian, such as one for the event \emph{Vacation} and the adverbial \emph{recently}, another one for the same event with the adverbial \emph{some time ago}, and so on. 

\subsection{Accuracy}
Accuracy measures how closely the model's predictions align with the empirical data from \citeA{kenneweg-etal-2024-empirical}, which serve as ground truth data. To evaluate the accuracy, we predict an adverbial's probability for a given elapsed time point of an event using our model and compute the average absolute difference between the predicted value and all votes from the empirical data. Table~\ref{tab:evaluation_accuracy} shows this average prediction error per event, adverbial and over all events and adverbials for our model and the normalized Gaussian.

\begin{table}[!ht]
\begin{center} 
\caption{Average error per elapsed time point, computed across events and adverbials for both our compositional model (Factorized Model) and normalized Gaussian (Non-factorized Model).} 
\label{tab:evaluation_accuracy} 
\vskip 0.12in
\begin{tabular}{ l l c c } 
\hline
Type &    Name        & \makecell{Factorized \\ Model} & \makecell{Non-facto- \\rized Model} \\
\hline
\multirow{6}{*}{Event} & Brushing Teeth   & 0.19 & 0.22 \\
                       & Birthday         & 0.19 & 0.19 \\
                       & Vacation         & 0.19 & 0.19 \\
                       & Sabbatical       & 0.21 & 0.22 \\
                       & Year Abroad      & 0.21 & 0.21 \\
                       & Marriage         & 0.19 & 0.18 \\
\hline
\multirow{4}{*}{Adverbial} & Just         & 0.16 & 0.15 \\
                           & Recently     & 0.18 & 0.17 \\
                           & Some time ago & 0.28 & 0.29 \\
                           & Long time ago & 0.18 & 0.20 \\
\hline
Overall           &      &   0.20 &   0.20 \\
\end{tabular} 
\end{center} 
\end{table}

\subsection{Extendability}
Extendability describes the ability of a system to adapt to novel scenarios or inputs. In our case, we are interested in understanding the effort involved in extending the model to new adverbials and/or events. The dataset used to fit the parameters of the model 
comprises 6 events and 4 vague temporal adverbials. For our compositional model, $6+4=10$ fitted functions were needed to include all events and vague temporal adverbials from the empirical dataset. The normalized Gaussian requires one function for each event-adverbial pair, resulting in $6*4=24$ functions. 

Expanding the compositional model to include one additional event would require only one additional function, yielding $7+4=11$ functions in total. The normalized Gaussian approach would require one additional function for each contained adverbial, yielding 4 additional functions and $7*4=28$ functions in total. Table~\ref{tab:evaluation_Extendability} illustrates this scalability effect by representing the number of required functions for up to 16 events and 16 adverbials.

\begin{table}[!ht]
\begin{center} 
\caption{Number of needed functions for x events and y adverbials for the compositional model (Factorized Model) and normalized Gaussian (Non-factorized Model).} 
\label{tab:evaluation_Extendability} 
\vskip 0.12in
\begin{tabular}{c c|c c} 
\hline
\multicolumn{2}{c|}{Number of} & \multicolumn{2}{c}{Needed Functions} \\
Events &  \multicolumn{1}{c|}{Adverbials} &\makecell{Factorized \\ Model} & \makecell{Non-factorized \\ Model} \\
\hline
2  &   2  & 4 & 4\\
2  &   4  & 6 & 8\\
2  &   8  & 10 & 16\\
2  &   16 & 18 & 32\\
4  &   16 & 20 & 64\\
8  &   16 & 24 & 128\\
16 &   16 & 32 & 256\\
\hline
\end{tabular} 
\end{center} 
\end{table}

% \begin{figure}
% \begin{center}
% \includegraphics[width=\linewidth]{Plots/needed_Functions.png}
% \end{center}
% \caption{Number of needed Functions for the compositional model and normalized Gaussian if we have x events and x adverbials.} 
% \label{plot:evaluation_needeFunctions}
% \end{figure}

\section{Discussion}

In developing this model, we have formally defined the \emph{reference class}—the context in which a vague predicate is interpreted—in line with Frege's principle of compositionality, by combining a vague temporal adverbial with a specific event. With this approach we extend previous probabilistic accounts of vagueness, which do not model the reference class compositionally in the Fregean sense \cite{DAMERAU1977, Heim, Kamp_Keenan_1975, Kennedy, Schmidt2010, Qing2014, Lassiter2017}.

Our initial analysis reveals that each examined event has its own distinct standard deviation (see Table~\ref{tab:fittingresults_event}). This suggests that every event possesses a \emph{characteristic temporal signature}, most likely determined by its frequency and duration \cite{jaarsveldschreuder}. For instance, \emph{brushing teeth} exhibits a relatively narrow standard deviation ($\sigma_e = 935$), whereas \emph{sabbatical} shows a considerably larger standard deviation ($\sigma_e = 798494$). These findings indicate that the \emph{characteristic temporal signature} shapes the reference class function (event function $P_{Ev}$); however, further investigation is needed to understand precisely how these \emph{characteristic temporal signature} exert their influence. Although the empirical dataset includes only 6 events - and more evaluation is necessary - our preliminary results (which match the accuracy of the baseline model; see Table~\ref{tab:evaluation_accuracy}) suggest that our compositional model effectively captures the meaning of vague temporal adverbials and is a good starting point for further research.

Following an argument rooted in Occam's Razor principle, our factorized model is clearly preferable to a non-factorized model as it has similar predictive power while representing a simpler explanation of the phenomenon. 

As shown in the `Evaluation' Section, its core strength lies in its extendability: rather than requiring a dedicated function for each event-adverbial pair, our compositional approach separates the modeling of events and adverbials into two distinct components. In doing so, the needed number of functions for the compositional model increases linearly as the number of events and adverbials grows. In contrast, for the baseline model (the normalized Gaussian), the number of functions would increase quadratically. Concretely (as shown in Table ~\ref{tab:evaluation_Extendability}), when modeling 16 events and 16 vague temporal adverbials, the normalized Gaussian would need $16*16=256$ functions, while our model requires $16+16=32$--only about 12.5\% of the baseline approach. 
%And this percentage gets even smaller when we include more adverbials and events: For 100 Events and 100 Adverbials our model would require 200 functions, the normalized Gaussian 10000, therefore our models needs only 2\% of the baseline approach. 
Despite this clear benefit in terms of extendability, the accuracy of our model is comparable to the accuracy of the baseline approach: both have an averaged error of 0.2 (see Table \ref{tab:evaluation_accuracy}).

While we have argued that our model is easier to extend compared to a non-factorized model, an inherent limitation of the model is that we would need to acquire additional data for new adverbials and events as in the study of \citeA{kenneweg-etal-2024-empirical}. An interesting direction for future work is whether the generalizability of the model can be improved by considering equivalence classes or categories of events that share a similar duration and frequency. 

%However, our model also has limitations, which lay in the used empirical dataset: \citeA{kenneweg-etal-2024-empirical} have collected only 100 votes for each elapsed time point of the 6 events and 4 vague temporal adverbials. Furthermore, we only knew that adding an additional event or adverbial to the models abilities requires one additional function but we do not provide an approach how we can decide the parameters of the adverbial function $P_{Adv}$ or event function $P_{Ev}$ without using an additional survey like the one in \citeA{kenneweg-etal-2024-empirical}.

\section{Conclusion \& Future Work}
We introduced a factorized compositional model that formally captures the semantics of vague temporal adverbials. Our approach models the application of a vague temporal adverbial to an event by defining two probability density functions: $P_{Adv}$ for the Adverbial and $P_{Ev}$ for the event. The resulting composition $P_{Adv}(P_{Ev}(t))$ computes the probability that an adverbial applies to an event that happened $t$ time units ago. In defining $P_{Ev}$, the reference class for the vague temporal adverbials, we extend previous approaches by formally defining a reference class of a vague predicate in line with Frege's principle of compositionality. 

By fitting our compositional model to empirical survey data from \citeA{kenneweg-etal-2024-empirical}, we instantiated the model for 6 events and 4 vague temporal adverbials. 

Our evaluation demonstrates that our compositional approach matches the predictive accuracy of a non-factorized baseline model while being simpler (less parameters) and easier to extend to new events and adverbials. However, it currently addresses only a limited range of 6 events and 4 vague temporal adverbials - far fewer than those encountered in everyday life and language. Although extending our model to additional events and adverbials requires only defining one function for each additional event or adverbial, further research is needed to find out how to systematically induce the functions' parameters, without collecting new data through surveys like the one from \citeA{kenneweg-etal-2024-empirical}. 

\bibliographystyle{apacite}

\setlength{\bibleftmargin}{.125in}
\setlength{\bibindent}{-\bibleftmargin}

\bibliography{CogSci_Template}

\end{document}